\documentclass[10pt,twocolumn]{article} 
\usepackage{simpleConference}
\usepackage{times}
\usepackage{graphicx}
\usepackage{amssymb}

\usepackage{algorithm}
\usepackage{algpseudocode}

\usepackage{tikz}
\usetikzlibrary{plotmarks}
\usetikzlibrary{intersections,shapes.arrows}
\usepackage{pgfplots}
\pgfplotsset{compat=1.12}
\usetikzlibrary{patterns, positioning, arrows,spy,shapes.geometric,calc}
\usetikzlibrary{positioning, fit, arrows.meta, shapes}
\usepackage{cite}
\usepackage{amsmath,amssymb}
\usepackage[pdftex,colorlinks,bookmarksnumbered,bookmarksopen,citecolor=red,urlcolor=red]{hyperref}
\usepackage[normalem]{ulem} 
\usepackage{breqn}
\usepackage{subfig}
\usepackage{epsfig}

\usepackage{amsfonts}
\usepackage{pgfplots}

\newcommand{\Lagr}{\mathcal{L}}

\newcommand{\bs}[1]{\boldsymbol{#1}}

\newcommand{\ddiff}[2]{\frac{\mathsf{D} #1}{\mathsf{D} #2}}

\begin{document}
\title{ A Thermodynamics-informed Active Learning Approach to Perception and Reasoning about Fluids}

\author{Beatriz Moya$^{1}$, Alberto Badias$^{2}$, David Gonzalez$^{1}$, Francisco Chinesta$^{3}$, Elias Cueto$^{1}$ \\
\\
$^{1}$Aragon institute in Engineering Research. University of Zaragoza, Spain \\
\{ beam, gonzal, ecueto\}@unizar.es\\
\\
$^{2}$Polytechnic University of Madrid, Spain \\
alberto.badias@upm.es\\
\\
$^{3}$ESI Group chair. PIMM Lab, ENSAM Institute of Technology, France \\
francisco.chinesta@ensam.eu \\
}

\maketitle
\thispagestyle{empty}

\begin{abstract}
Learning and reasoning about physical phenomena is still a challenge in robotics development, and computational sciences play a capital role in the search for accurate methods able to provide explanations for past events and rigorous forecasts of future situations. We propose a thermodynamics-informed active learning strategy for fluid perception and reasoning from observations. As a model problem, we take the sloshing phenomena of different fluids contained in a glass. Starting from full-field and high-resolution synthetic data for a particular fluid, we develop a method for the tracking (perception) and analysis (reasoning) of any previously unseen liquid whose free surface is observed with a commodity camera. This approach demonstrates the importance of physics and knowledge not only in data-driven (grey box) modeling but also in the correction for real physics adaptation in low data regimes and partial observations of the dynamics. The method presented is extensible to other domains such as the development of cognitive digital twins, able to learn from observation of phenomena for which they have not been trained explicitly.

\end{abstract}

\section{Introduction}
The research community has witnessed great advances in deep learning and Artificial Intelligence (AI) concerning the imitation of human-like skills \cite{liu2021role} \cite{schenck2018perceiving}. However, despite experiencing great flourishing, works are still transitioning to the development of the so-called Artificial General Intelligence (AGI) \cite{goertzel2007artificial}. In this next step of AI, we approach more anthropomorphic skills, independence and high-level reasoning for decision-making \cite{zhang2021generalized} and behavior learning \cite{andrychowicz2020learning} \cite{levine2016end}. One of the requirements to cross the boundary between AI to AGI is to improve the development of sensory and physics perception in machine systems. Physics perception refers to the interpretation and reasoning skills that interpret sensed data. These technologies play a capital role in robotics, especially for planning and control tasks \cite{schenck2018perceiving}. To develop systems able to reason about their surrounding environment we need interactive, real-time simulators of the real-world physics, constructed by a judicious mixing of theory, data, and computation \cite{liu2021role}. 

In some sense, these systems resemble the now popular digital twins. However, these new systems, able to perceive and reasoning about physics, are more aligned with the concept of cognitive or hybrid digital twin \cite{chinesta2020virtual}. In these, simulation is enriched with new measurements to correct the biased predictions that result from comparing the output of the simulation with the ground truth, and to optimize the accuracy of the solution in new scenarios that the user does not control. This framework arises as a solution to complement current techniques in the so-called smart data paradigm, to profit from data and perform the desired update, adaptation, and knowledge enrichment promoting the efficient use of data. 

This can be seen, alternatively, also as a particular case of Dynamic Data Driven Applications Systems (DDDAS) \cite{darema2013dynamic} \cite{darema2005grid}. In this context, data is introduced dynamically in a continuous learning loop to improve the performance of the obtained forecasting. Data-driven physics-informed simulators reach higher generalization than unconstrained models or purely theoretical approximations, merging the knowledge coming from data distributions and well-known physics priors. However, there are still difficulties to match the model outputs with an evolving real physical environment \cite{atkeson1997learning}. For this reason, the learned simulator can be formulated as a hybrid twin, or more generally, a DDDAS, to adapt to new scenarios through continuous observation and measurement of labeled data. 

Blakseth et al. propose the system \textit{CoSTA} \cite{blakseth2022deep}. They develop a hybrid strategy to complement a physics-based model with a second, data-driven, term that will be in charge of learning the corrections. In \cite{moya2020digital}, the authors proposed a hybrid twin of a hyperelastic beam with moving loads, displayed by means of augmented reality. In the field of perception and reasoning, \cite{blakseth2022deep} exploits DDDAs in scene-understanding in unknown scenarios, where the system provides classification or enriches the knowledge previously acquired depending on its level of confidence. \cite{schenck2017reasoning} proposes a system of physical intuition about liquids that is corrected from observations but without restrictions of physical priors. 

Reinforcement learning (RL) also constitutes an interesting framework for the construction of perception strategies from the interaction with the medium  \cite{sutton2018reinforcement}. Its main purpose is to reward those actions that lead to a bound objective. Reinforcement learning can also be applied to the optimization of learned simulators of complex dynamical systems, thus mimicking the way we, humans, learn intuitive representations of the physical world that surrounds us, to interact with this environment. Examples of model control for adaptation with reinforcement learning include the Burger's Equation as a common benchmark \cite{benosman2021reinforcement} \cite{bassenne2019computational} \cite{wang2019learning}, or the Kuramoto-Sivashinsky Equation \cite{bucci2019control}. In this context, this strategy also targets turbulence and flow control problems, where the object of the optimization is also the physics simulator \cite{rabault2019accelerating} \cite{ren2021applying} \cite{verma2018efficient} \cite{garnier2021review} \cite{novati2021automating} \cite{rabault2019artificial} \cite{ren2021applying}.

In this work we take a sloshing fluid in a glass as the model problem. Despite its apparent simplicity, the fluid is highly deformable (something always difficult for perception systems), highly non-linear, and dissipative (conservative problems have been demonstrated to be much easier to learn).  We develop a method for the perception (tracking of the free surface of the fluid) and reasoning (providing the user with full-field information---velocity, stress, ...) about the physical state of a sloshing fluid. While previous works employ simulation as the engine of physical scene understanding (see \cite{allen2020rapid} or \cite{battaglia2013simulation}, for instance), thus needing previous knowledge of the physics of the scene and a pre-defined simulator, our method constructs on the fly a learned simulator. This learned simulator is built by resorting initially to synthetic full-field data, possibly coming from different fluids. At runtime, when faced to video streams of a previously unseen surrounding reality, it makes use of an active learning strategy in a thermodynamics-informed setting to correct systematic deviations of the observed reality from its predictions.  Our approach ensures the compliance to first principles---conservation of energy, non-negative entropy production---of the resulting simulations, even if they are constructed from partial observations of the reality (in our case, the observation of the free surface of the fluid).

Both the originally learned simulator and the active learning strategy are based on the leverage of the General Equation for the Non-Equilibrium Reversible-Irreversible Coupling (GENERIC) formalism as an inductive bias to guide the training by the laws of the thermodynamics \cite{grmela1997dynamics}. In addition, the imposition of inductive biases could be found advantageous in the correction process to reduce error bounds and adapt to new scenarios more efficiently \cite{gonzalez2019learning}. The physical knowledge already gained in the source model is to be preserved for the sake of the physical interpretability of the results and the success of the adaptation. The proposed thermodynamics-informed scheme will adapt to the inherent dissipative nature of real phenomena. The model is to be corrected from the sole observation of the free surface of the new liquid if biased deviations are noticed. As a result, we obtain a corrected, augmented intelligence system that performs the integration of the fluid dynamics in real-time from the evaluation of the free surface evolution. The output is the new state of the fluid, as the result of its integration in time. This approach is coupled with a computer vision system to build a closed-loop algorithm that learns liquid slosh from real measurements. The adaptation is performed in an offline phase to reward the cumulative improvement over online acquired data. After the correction, the simulation of the dynamics is accomplished in real-time, obtaining improved predictions of the phenomena taking place in the physical world. 

The development of this method revolves around the convergence of the aforementioned DDDAS, active learning, and transfer learning strategies. Transfer learning profits from a model already learned to apply the gained knowledge to new tasks (in our case, previously unseen fluids). This technique has also been considered in the a-posteriori correction of models such as manifold learning in reduce order modeling of fluids \cite{mohebujjaman2019physically} and dynamical optimization \cite{laroche2017transfer} \cite{goswami2020transfer} \cite{guastoni2021convolutional}.

\section{Problem formulation}

We describe the evolution of a complex physical system as a function of a set of phase variables $\bs s_n=\bs s(n\Delta t)$, where $t$ represents the time, that defines its state. Both the source model and the correction algorithms are learned by employing the GENERIC equation as an inductive bias \cite{grmela1997dynamics}. GENERIC proposes a mesoscopic formulation of dynamical systems to describe their behavior in terms of the evolution of energy and entropy in the system. This approximation for dynamical systems can be discretized in time and its constituting terms can be thus inferred from data \cite{gonzalez2019thermodynamically}. The so-called Structure-Preserving Neural Networks (SPNN) \cite{hernandez2021structure} \cite{hernandez2021deep} are deep learning architectures that employ the GENERIC formalism as an inductive bias to learn dynamical patterns from data and perform the simulations with guaranteed physically meaningful results.

Our method constructs in a first step a learned simulator, by employing a neural network trained offline with synthetic data coming from computational simulations \cite{moya2021physics}. This level of description agrees with theories that have demonstrated that human reasoning about fluid physics can be understood as a learned simulator operating at a coarse-grained level of description \cite{bates2019modeling}. To train this first simulator, we employ data coming from full-field and high-fidelity smooth particle hydrodynamics of a source fluid---in our case, glycerine. We apply different initial velocities to the glass to trigger the slosh and build an initial database.

We employ the same glass for all the experiments, and we fill it with a liquid up to the same level for all the experiments. Hence, neither the geometry of the container nor the liquid volume are considered as a variable for our problem. As a result, since the fluid volume is the same, we start from the same SPH discretization in each simulation. This assumption is employed to justify the use of fully connected neural networks in this formulation. 

\begin{figure*}[!t]
\centering
\subfloat{\includegraphics[width=0.8\linewidth]{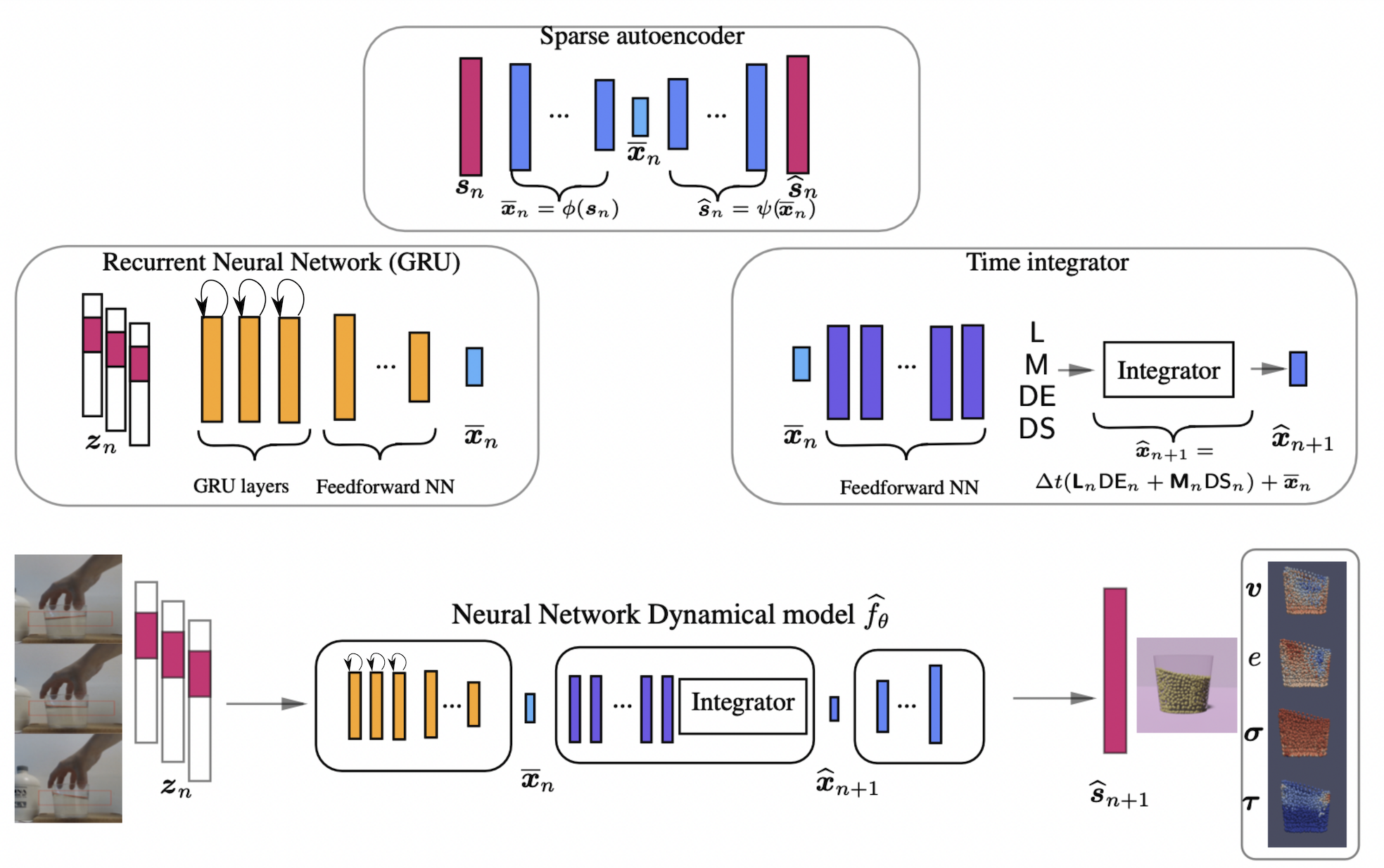}}%
\hfil
\caption{Initial learned simulator. It is composed of three parts. From synthetic, full-field data, we first train off-line an autoencoder to find a low dimensional manifold of the dynamics under study. (Left) There we train a structure-preserving network to determine its constituents: the symplectic matrix $ \bs L$, the dissipation matrix $\bs M$, an approximation to the gradient of energy, $DE$ and an approximation to the gradient of entropy, $DS$. (Right) Since we have online access to the free surface of the fluid only (pink entries $\bs z_n$ in the input vectors $\bs s_n$), a recurrent neural network exploits knowledge in sequences of data to find the embedding to the low dimensional manifold discovered off-line from full-field data (transfer learning). The Structure-Preserving Neural Network (SPNN) propagates in time the evolution of the dynamics in the manifold, fulfilling the principles of thermodynamics as a major requirement. Finally, the decoder of the autoencoder projects the new state of the fluid to the world coordinates to recover the volume of the liquid and other quantities of interest ($\bs v$ velocity, $e$ energy, $\bs \sigma$ normal stress, $ \bs \tau$ shear stress, grouped in the full-field vector $\bs s_{n+1}$), whose knowledge is employed in the integration in the latent manifold.}
\label{fig1}
\end{figure*}

This initial simulator is first trained off-line with synthetic data coming from the simulation of glycerine sloshing phenomena (viscosity $\mu = 0.950Ns/m^2$ and density $\rho = 1261kg/m^3$) \cite{gregory2018physical}. This learned simulator unveils the intrinsic dimensionality of data using a sparse autoencoder, to develop a reduced-order model, able to run under stringent real-time constraints \cite{goodfellow2016deep}. Fig.~\ref{fig1} describes the architecture of the original simulator. In this way, we compress the information within the initial dataset of full-field state vectors, which consist of the position, velocity, energy, and stress tensor of all the particles of the SPH discretization. This knowledge is transferred to two new networks. Then we train the time integrator of the evolution of the dynamics in the latent manifold based on structure-preserving NN. 

Our learned simulator is then faced with online measurements performed by a commodity camera. During this online phase, we have access to the position of the free surface of the fluid only. We, therefore, substitute the encoder with a recurrent neural network to connect these partial measurements of the free surface of the liquid to the already unveiled latent manifold of the dynamics. We assemble these three networks to develop the simulation-in-the-loop system that, coupled with computer vision elements, will correct the results from new data acquired from observations. 

\section{Method}

\subsection{Data acquisition}

\subsubsection{Computational datasets}

We first conduct a set of purely computational experiments. This allows us to obtain accurate error measurements against high-fidelity simulations that are considered as the ground truth. Synthetic data are obtained from computational simulations using the software Abaqus (Dassault Syst{\`e}mes, Simulia Corp).

The method has been tested against a total of ten different liquids, presenting both Newtonian and non-Newtonian behavior.  We first train a learned simulator for glycerine \cite{gregory2018physical}. With this simulator, we try to understand the behavior of water and melted butter, for instance, both described as Newtonian fluids. Although blood is sometimes considered to be Newtonian, it has been described as non-Newtonian \cite{liquids}. It is considered to be shear thinning, becoming less viscous under the stress applied. The general description of the rheology and change of properties of fluids is performed based on the Herschel-Bulkley model \cite{herschel1926konsistenzmessungen}
$$
\bs \tau(t)=k \bs \gamma^n(t)+\bs \tau_0,
$$
where $\bs \tau$ is the shear stress, $\bs\tau_0$ the yield stress, $\bs \gamma$ the shear rate, $k$ the consistency index and $n$ the flow index. For shear-thinning fluids, $\bs \tau_0$ is greater than 0, and the flow index is $n>1$.  In the proposed case of computational blood, the computational liquid is defined by the constants $k=0.017 Pa s$, $n=0.708$, and $\bs \tau_0=0$. 

These datasets are publicly available at \url{https://github.com/beatrizmoya/sloshingfluids/}.

\subsubsection{Real world dataset}

{ We then employ a stereo camera for data acquisition of the free surface of real-world liquids. Although there are known applications of sophisticated tools such as PIV cameras \cite{rabault2017performing} \cite{cai2019particle}, we suppose only have access to the position of the free surface with an ordinary stereo camera. The model chosen is a Real-Sense D435 \url{https://www.intelrealsense.com/depth-camera-d435/}. We deliberately omit the possibility of using PIV systems, for instance, to force our system to work in a partial data regime, i.e., without any velocity data other than the reconstruction of the free surface.}

In a camera model, the correlation of the pixels in a picture, or frame, with the point they represent in the real world $ \bs p_w$ is given by the extrinsic and the intrinsic parameters. The extrinsic parameters, rotation $\bs R$ and translation $\bs t$ matrices define the position and orientation of the camera referring to the world frame. It outputs the relative position of the camera in the real world. On the other hand, the intrinsic parameters $\bs K$ map the pixel coordinates to the camera coordinates. The camera provides the intrinsic and extrinsic parameters to convert the pixel coordinates on the image directly to three-dimensional coordinates they represent in the real world,
\[
\begin{bmatrix}
  u\\ 
  v\\
  1   
\end{bmatrix}
=
\begin{bmatrix}
  f_{x} & 0 & c_{x}\\ 
  0 & f_{y} & c_{y}\\
  0 & 0 & 1   
\end{bmatrix}
\begin{bmatrix}
  r_{11} & r_{12}  & r_{13}& t_{1} \\ 
  r_{21}  & r_{22} & r_{23}& t_{2}\\
  r_{31} & r_{32} & r_{33} & t_{3}
\end{bmatrix} 
\begin{bmatrix}
  X\\ 
  Y\\
  Z\\
  1   
\end{bmatrix},
\]
or, in brief,
$$
\tilde{\bs x}_{s}=\bs K[\bs R|\bs t] \bs p_{w}.
$$

The camera has stereoscopic depth technology to perceive the depth field of every pixel in the frame. With two lenses, triangulation \cite{andrew2001multiple} gives the 3D position of the points. 

Despite the accuracy of the camera, fine-tuning was necessary due to the lack of texture of the glass and liquids. These conditions difficult the tracking of points, resulting in errors. The application of filters (hole-filling and edge-preserving filters) improves the performance of depth estimation. In optimal conditions, the error in depth reconstruction is lower than $2\%$ in a $0-2 m$ range. After the fine-tuning of the camera, in the threshold where we locate the glass for the recording, the camera noise in depth estimation is of the order of two millimeters. 

\begin{figure*}[!t]
\centering
\subfloat{\includegraphics[width=0.7\linewidth]{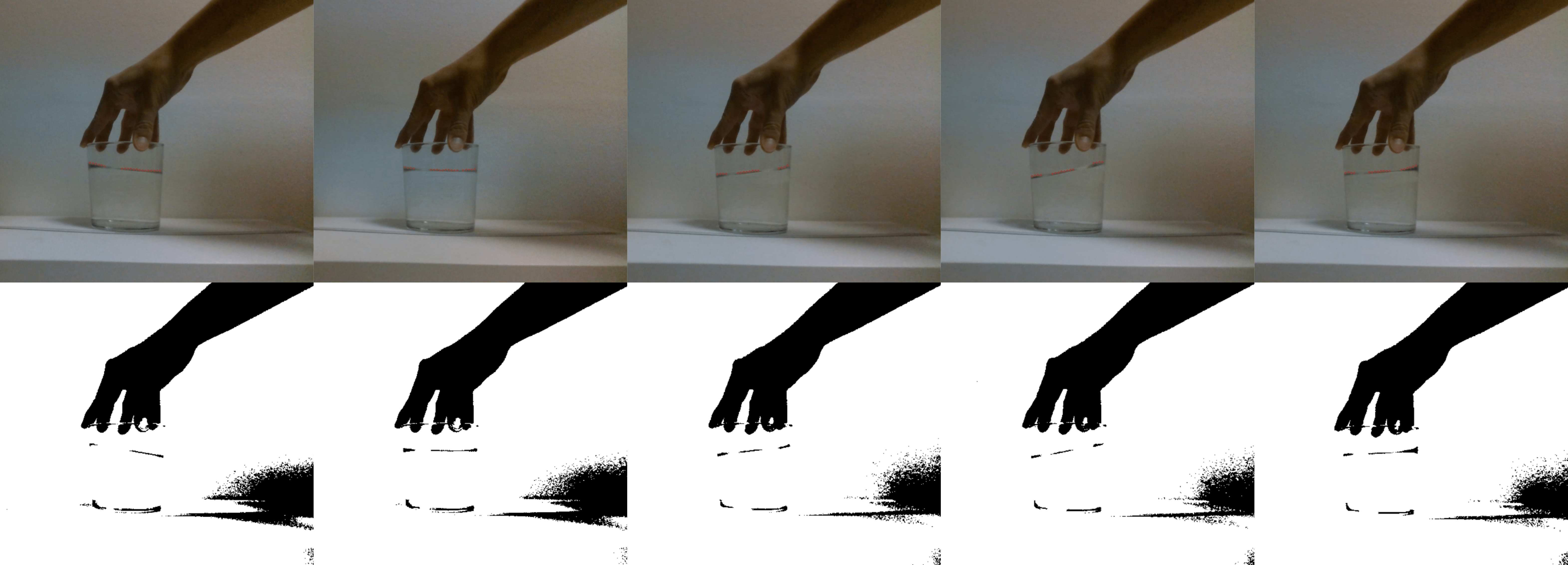}}
\hfil
\caption{Representation of surface tracking. Each color frame is converted to a binary frame where the surface is detected (shown in red in the correspondent color frame). The depth map is built upon these points. For transparent liquids, like the glass of water shown in the picture, the depth map is of lower quality resulting in an incomplete detection of the surface.}
\label{fig2}
\end{figure*}

Given a sufficient quality of depth resolution, we perform the observation and tracking of the free surface. We extract the profile of the free surface from binary (black and white) images. Under appropriate adjustment, this representation shows a black to white gradient between the liquid and the environment, see Fig. \ref{fig2}. These color changes enable to track the free surface. We extract the depth maps of the pixels of the aforementioned boundary. No further smoothing of the data is performed. Realtime tracking is performed with a camera resolution of $480 \times 640$ pixels. 

From the acquired map, we project the pixel coordinates of the free surface to real-world coordinates and select their horizontal and vertical coordinates, or 2D real coordinates. Given the density of points detected on the free surface, we perform a simple linear interpolation to output the vertical displacement at equally spaced points of the free surface. This state information of the free surface is assembled in sequences to feed the source algorithm. 

The dataset of the free-surface measurements performed on real liquids has been made publicly available: \url{https://github.com/beatrizmoya/RLfluidperception}

\subsection{Physics-constrained learning}

Once data are captured, either in computational, synthetic form or as a recording of the physical reality, our aim is to identify the state of the observed scene (perception) in terms of observable and non-observable variables---such as stresses, for instance---and to predict future states of the system (reasoning) by learned simulation. In the last years, a growing interest has been noticed in the incorporation of known physics in the form of inductive biases to this learning procedure. These representations will be essential for the generalization and interpretability of the method. Of particular importance, when some form of conservation (related to symmetries through Noether's theorem) is present, this can be imposed to the learning procedure by invoking Lagrangian or Hamiltonian frameworks. Despite their success, many dynamical systems are also compromised by dissipative effects, which implies that this framework of techniques is no longer valid for application. This premise also matches the dissipative nature of the real world and the entropy that emerges from the lack of information. Fluid dynamics, as well as other dissipative dynamical systems, own a so-called metriplectic structure \cite{kraus2021metriplectic}. This formalism is suitable for cases in which the conservative Hamiltonian description of a dynamical system includes unresolved degrees of freedom, that are not included in mesoscopic descriptions and that introduce dissipation by the fluctuation-dissipation theorem \cite{kubo1966fluctuation}. Consider an initial microscale description at the molecular dynamics scale, that can be expressed in terms of a purely conservative formulation. The degrees of freedom, and therefore knowledge, that we omit growing from the micro to the meso and macro scales introduce dissipation that is included in the metric part of the formulation.

This structure-preserving formulation guarantees the conservation of critical quantities (mass, momentum), and the thermodynamical admissibility of the evolution of the system under study. GENERIC is the chosen thermodynamic framework to derive a structure-preserving time integrator for describing the time evolution of a system concerning the evolution of its energy and entropy functionals \cite{grmela1997dynamics}. As a consequence, this framework is also valid for dynamical systems that go beyond equilibrium in a thermodynamical context. The GENERIC formulation is usually written as follows:
$$
\frac{\mathrm{d} \bs s}{\mathrm{d} t} = \bs L\nabla E +\bs M \nabla S.
$$
Here, $\bs s$ denotes a set of independent state variables that fully describe the thermodynamical state of the fluid. Without that information, we lack a GENERIC structure. Fluid dynamics are fully described in terms of the position and momentum of the particle discretization, internal energy, and, in the case of learning more complex fluids, the extra-stress tensor related to their microscopic evolution \cite{espanol1999thermodynamically}. $E(\bs s)$ and $S(\bs s)$ are the global energy and entropy of the system. $\bs L(\bs s)$ is the Poisson matrix. It is skew-symmetric, and together with the gradient of energy $\nabla E$, characterizes the reversible part of the studied dynamics. $\bs M(\bs s)$ is the friction matrix, which describes the dissipative irreversible characteristics of the system in conjunction with the entropy gradient $\nabla S$. $\bs M$ is symmetric positive semidefinite. 

On top of the aforementioned formulation and matrix descriptions, we must also ensure the fulfillment of the so-called \textit{degeneracy conditions}:
$$
\bs L \frac{\partial S}{\partial \bs s} = \bs M \frac{\partial E}{\partial \bs s}=\bs 0,
$$
that guarantee that the energy $E$ is not involved in the entropy production of the dissipative part of the dynamics, and that the entropy $S$ does not contribute to the energy conservation. 

The GENERIC equation can be discretized in time by a forward Euler scheme formulated in time increments $\Delta t$, and the resulting formulation subsequently inferred from data:
$$
\bs s_{n+1}=\bs s_{n} + \Delta t \left (\boldsymbol{\mathsf{L}}_n\frac{\mathsf{D} E_n}{\mathsf{D} \bs s_{n}} +\boldsymbol{\mathsf{M}}_n \frac{\mathsf{D}S_n}{\mathsf{D}\bs s_{n}}  \right ), 
$$
subjected to the degeneracy conditions:
\begin{equation}\label{eq:degendisc}
\boldsymbol{\mathsf{L}}_n \ddiff{S_n}{\bs{x}_n} =\bs  0, \quad
\boldsymbol{\mathsf{M}}_n \ddiff{E_n}{\bs{x}_n} = \bs 0,
\end{equation}
ensuring the thermodynamical consistency of the resulting model.

Structure-preserving neural networks (SPNN) embed GENERIC into a deep learning architecture to unveil the value of the main elements of the equation \cite{hernandez2021deep} \cite{hernandez2021structure}. In this framework, the discretized gradients of energy and entropy are the targets of the optimization of the net. In addition, although $ \boldsymbol{\mathsf{L}}$ and $ \boldsymbol{\mathsf{M}}$ are usually known in the literature, they are unknown in the low dimensional manifold where we simulate the dynamics. Hence, the network learns $\boldsymbol{\mathsf{L}}, \boldsymbol{\mathsf{M}}, \frac{\mathsf{D} E_n}{\mathsf{D} \bs s_{n}}, \frac{\mathsf{D} S_n}{\mathsf{D} \bs s_{n}} $ from data. This is represented in Fig. \ref{fig1}. 

\subsection{Correction algorithm}

As already stated, our system is initially trained to perceive and reason about glycerine. But it is then faced to different, previously unseen fluids. To be able to provide with useful and credible predictions for these new fluids, an active learning strategy is employed. 

\begin{figure}[h!]
\centering
\includegraphics[width=\columnwidth]{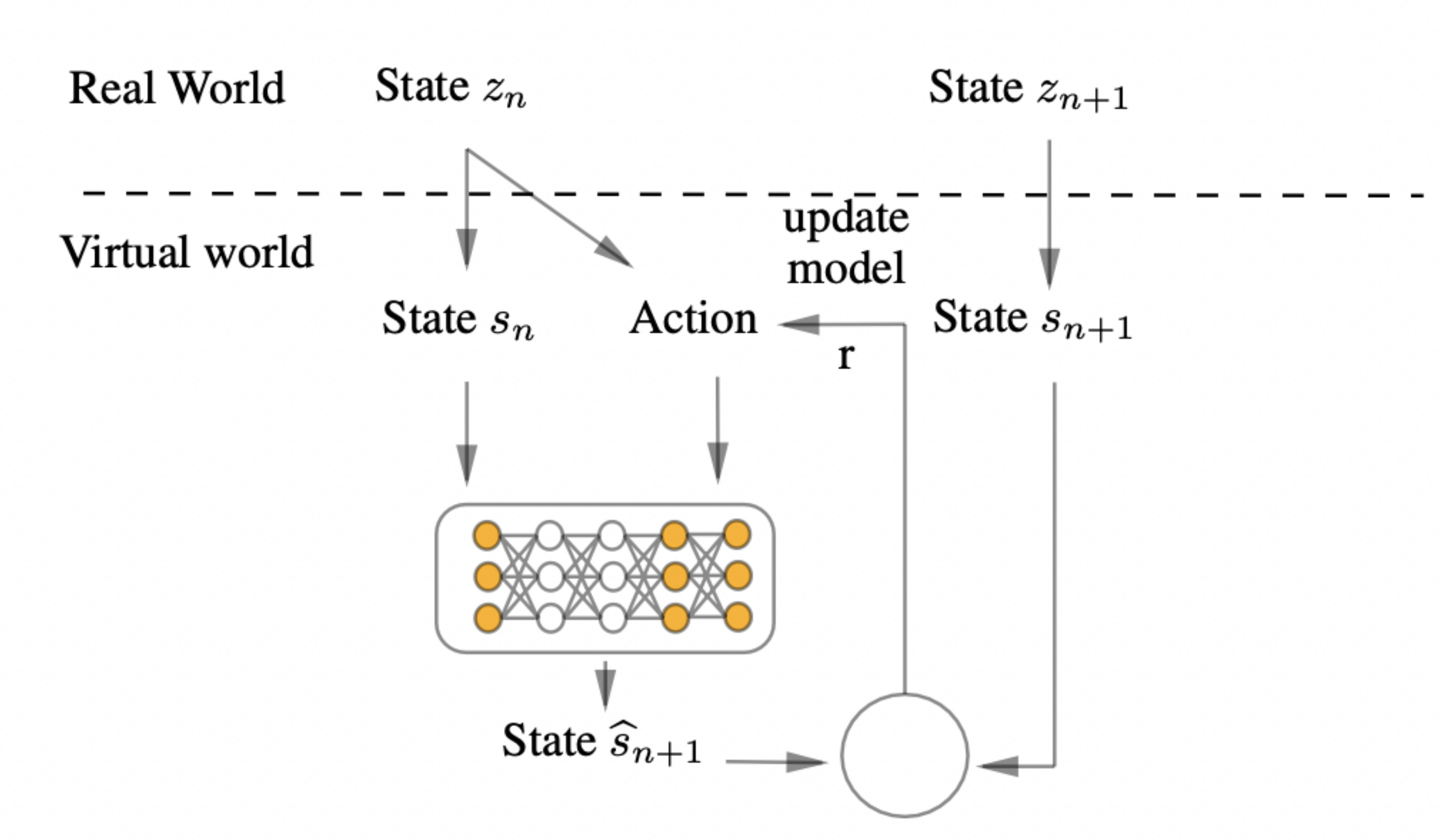}
\caption{Update scheme. Given the partial observation of the free surface $\bs z_n$ at time $t = n \times \Delta t$ our method estimates the full state of the fluid $  \hat{\bs s}_{n+1}$. Its similarity with the next state $\bs s_{n+1}$ is evaluated to update the model. To preserve the patterns previously learnt, only a few layers are activated (in yellow) for backpropagation of the error and correction of the model.}
\label{fig3}
\end{figure}

{ The sloshing dynamics are described by its dynamical state $\bs s_n \in \mathcal{S}$, where $\mathcal{S}$ represents the state space where the dynamics are embedded. From that state $\bs s_n$, we actually have access to a partial observation $\bs z_n\subset \bs s_n$. If $\bs s_n$ refers to the fully dynamical description of the slosh, $\bs z_n$ stands for the observed degrees of freedom on the free surface. Free surface detection and tracking are performed employing computer vision techniques. The slosh evolves in time, and the integrator provides results for future states $\bs s_{t+1}$, not only for the observed ones, $\bs z_{t+1}$. For its calculation, a fully connected neural network based on the GENERIC formalism estimates the parameters of the equation. Mimicking reinforcement learning strategies, the ultimate objective of this work is to find the most convenient policy $\pi_{\theta}(a|\bs s)$ so that we obtain a satisfactory estimation of the parameters of GENERIC, that results in an accurate reconstruction of the motion of the liquid.  $\pi_{\theta}(a|\bs s)$ depends on the relationship between the input sequence and the prediction, which includes the recurrent neural networks, the learned simulator and the selection of $\boldsymbol{\mathsf{L}}, \boldsymbol{\mathsf{M}}, \frac{\mathsf{D} E_n}{\mathsf{D} \bs s_{n}}, \frac{\mathsf{D} S_n}{\mathsf{D} \bs s_{n}}$ , and a decoder that reconstructs the full-field description of the liquid $\bs s_{n+1}$. $\theta$ represents, as usual, the hyperparameters of the network.}

The loss is related to the accuracy of the free surface motion reconstruction. Through this, we reward the improvements of the network that lead to a closer result to the ground truth. We choose an $L_2$ norm for this purpose. We also reward those solutions that fulfill the degeneracy conditions established in the GENERIC formalism. As a result, the reward function is composed of these two terms, adequately weighted to control the influence of each term in the optimization of the network. The degeneracy condition is thus considered as a soft constraint:
\begin{multline} \label{reward}
r_{\pi} =\lambda \frac{1}{N}\sum_{N}\left \| \bs z_{n+1}-\hat{\bs z}_{n+1}  \right \|^2   \\
+ \frac{1}{N}\sum_{N}\left(\left \| \boldsymbol{\mathsf{L}}_n \frac{\mathsf{D} E_n}{\mathsf{D} \bs s_{n}} \right \|^2 + \left \| \boldsymbol{\mathsf{M}}_n \frac{\mathsf{D} S_n}{\mathsf{D} \bs s_{n}} \right \|^2\right).
\end{multline}
At each loop of the training, the algorithm is fed with $N$ data samples coming from the interaction with the environment. With them, the network parameters are updated. This process is repeated until we reach convergence in the adaptation process.

\begin{algorithm}
{ 
\caption{Update pseudocode}\label{alg:cap}
\begin{algorithmic}
\Require Free surface information $\bs z_n \in Z$ as a subset of the full state $\bs s_n \in S$, and the source model $\pi_{\theta}$
\Ensure Next dynamical states in time $\bs s_{n+1}$

\For{Iterations until convergence}
	\For{n=1 to N sequences}
		\State Encoder $\overline{\bs x}_n \gets \phi(\bs z_n)$;
         	\State Compute forward propagation and determine action $a_n \gets \pi_{\theta}(\overline{\bs x}_n )$, with $a_n=\left [\boldsymbol{\mathsf{L}}_n, \boldsymbol{\mathsf{M}}_n, {\mathsf{DE}}_n,{\mathsf{DS}}_n \right ]$;
		 \State Determine next integration step $\widehat{ \bs x}_{n+1}\gets \Delta t(\boldsymbol{\mathsf{L}}_{n} \mathsf{DE}_n+\boldsymbol{\mathsf{M}}_{n} \mathsf{DS}_{n})+\overline{\bs x}_n$;
		 \State Decoder $\widehat{\bs s}_{n+1} \gets \psi(\widehat{\bs x}_{n+1})$;
	\EndFor
	\State Extract free surface of $\widehat{\bs s}_{n+1}$;
	\State Compute reward $r_{\pi}$;
	\State Backward propagation and update behavior of $\pi_{\theta}$;
\EndFor
\\
\Return Optimized perception and reasoning approximation $\pi_{\theta}$
\end{algorithmic}
}
\end{algorithm}

As described in the algorithm, we first collect information coming from the observation of a fluid. Then, we compute the actions to obtain $\boldsymbol{\mathsf{L}}, \boldsymbol{\mathsf{M}}, \frac{\mathsf{D} S_n}{\mathsf{D} \bs s_{n}}, \frac{\mathsf{D} E_n}{\mathsf{D} \bs s_{n}}$ and perform the time integration for each given state. We then compute the loss with this information, and perform backpropagation to update the neural network. 

The backpropagation is computed through selected layers of the whole source model. This approach, in conjunction with low learning rates, ensure the minimum loss of information from the previous model to preserve the insights already learned with GENERIC. We profit from these known patterns to deal with the limitations coming from partial observations and low data regimes.

\section{Initial learned simulator}

The base simulator for glycerine is trained with synthetic data coming from the simulation of a fluid discretized in particles based on the smooth particle hydrodynamics technique. This coarse level of description is appropriate to have sufficient information about the dynamics to train learned simulators without compromising time. The state of the fluid is described by a set of state variables (position, velocity, energy, and stress tensor of each particle), evaluated at each particle of the discretization.

Due to the complexity of the problem, the simulator involves three different steps, and transfer learning is applied to carry the knowledge from simpler architectures to more elaborated ones. First, we train a fully-connected autoencoder to embed the data in a low-dimensional manifold where we will perform the integration in time. It consists of two parts: the encoder and the decoder. The encoder learns a mapping $\phi: \mathbb{R}^D  \to  \mathbb{R}^d$ to a low dimensional manifold where the dynamics are embedded. The decoder  $\psi: \mathbb{R}^d  \to  \mathbb{R}^D$ has the same structure of the encoder, but inverted, to project state vectors of the latent space to the full order space. Not only the computational cost for training the integrator will be reduced, but also the pre-processing will facilitate the learning process. By applying this model order reduction we are already triggering the emergence of the patterns, and encouraging the net to learn the main features of data. We train five autoencoders in parallel, one per each group of the already mentioned state variables, to capture the intrinsic insight of each one. The final latent manifold results from merging the latent subspaces of each autoencoder. The bottleneck of each autoencoder is truncated to reduce the dimensionality by forcing sparsity on its learning. Thus, for a given bottleneck, only a few latent values have a resemblant order of magnitude greater than zero. The decoder mirrors the encoder. The resulting latent manifold has 13 dimensions. The decoder is part of the new model, but it is not considered in the correction. Sparsity is imposed with a $L_1$ regularizer that penalizes activation in the bottleneck to enforce a higher reduction. This penalization complements the mean squared error reconstruction loss of the solution of the autoencoder: 

$$
\mathcal{L}_{AE}=\frac{1}{N}\sum_{i=1}^{N}(\bs s_i-\hat{\bs s}_i)^2 + \lambda_{\text{reg}} \sum_{i=1}^{N} \left | \bs x_i \right |. 
$$

The sparsity loss is weighted by a lambda factor to control its influence on the learning process. In the source model, the weight factor is defined within the range $\lambda_{\text{reg}}= 0.001-0.005$. As we train an autoencoder for each group of state variables, it is chosen according to each case.

Since the full-field state of the fluid cannot be evaluated, i.e., we cannot measure with a camera the internal energy of points of the fluid, we train a mapping from the available information of the free surface to the latent manifold in a second step of the procedure. Hence, we transfer the information from the latent manifold previously obtained to a recurrent neural network to substitute the encoder. The recurrent neural network exploits the analysis of sequences of data to distill the insights about the dynamics that enable finding a mapping between the observations and the latent manifold, which contains the full information about the dynamics embedded. We employ a gated recurrent unit (GRU) structure \cite{cho2014learning}. This architecture will be trained with the decoder frozen. We need a sequence of at least 16 snapshots to correlate the measurements of the free surface with the latent manifold. The net consists of three GRU hidden layers of 26 neurons, and one last feed-forward fully connected layer to connect the last GRU layer to the latent space of size 13. The sequence used contains the snapshot at time $t$, and the 15 previous snapshots. The network learns a mapping from each sequence of snapshots to the representation of the state of the snapshot at time $t$ in the latent manifold. We evaluate the accuracy of the mapping by measuring the discrepancy between the output of the GRU, the predicted reduced-order representation, and the ground truth latent vector:
$$
\Lagr_{\text{GRU}}=  \frac{1}{N_{\text{snap}}}\sum_{i=1}^{N_{\text{snap}}}(\bs x_i-\hat{\bs x}_i)^2.
$$

The architecture designed for the GRU was tested also with vanilla RNN and LSTM networks. While ordinary RNN did not succeed at finding a mapping, LSTM reached a performance similar to the GRU. However, we decided to use the GRU for simplicity. 

Finally, the structure-preserving neural network is trained for time integration of the dynamics based on the reconstruction accuracy of the solution in the latent manifold, as well as the compliance of the degeneracy conditions. This network is composed of 13 hidden layers of size 195 with ReLU activations but for the first and last layers, which have linear activations.  The loss $\Lagr_{\text{SPNN}}$ used for the training includes the mean squared error reconstruction error of the prediction $\Lagr^{\text{mse}}_{\text{SPNN}}$ and the loss related to the degeneracy conditions $\Lagr^{\text{deg}}_{\text{SPNN}}$. The reconstruction error is weighted by a $\lambda$ factor to prioritize this term. In this model, $\lambda_{\text{SPNN}} = 10^3$. 

$$
\Lagr^{\text{mse}}_{\text{SPNN}} =  \frac{1}{N_{\text{snap}}}\sum_{i=1}^{N_{\text{snap}}}(x_i-\hat{x}_i)^2,
$$

$$
\Lagr^{\text{deg}}_{\text{SPNN}} =  \frac{1}{N_{\text{snap}}}\sum_{i=1}^{N_{\text{snap}}}(\mathsf L_i \mathsf {DS}_i)^2+(\mathsf M_i \mathsf {DE}_i)^2,
$$

$$
\Lagr_{\text{SPNN}} =  \lambda^{\text{mse}}_{\text{SPNN}} \Lagr^{\text{mse}}_{\text{SPNN}} +\Lagr_{\text{deg}}^{\text{SPNN}}.
$$

{ The parameters used for reproducibility are shown in table \ref{table_param}.}

\begin{table*}[]
{ 
\begin{tabular}{@{}llllllll@{}}
\hline
                     & lr   & wd   & Hidden Layers                          & Input size & Output size & $\lambda$ & epochs \\ \hline
Autoencoder $q$      & 1e-4 & 1e-6 & N=2 of 120 neurons                     & 6402       & 20          & 2000      & 10000  \\
Autoencoder $v$      & 1e-4 & 1e-5 & N=4 of 200 neurons                     & 6402       & 20          & 2000      & 10000  \\
Autoencoder $e$      & 1e-4 & 1e-5 & N=3 of 40 neurons                      & 2134       & 10          & 2000      & 10000  \\
Autoencoder $\sigma$ & 1e-4 & 1e-5 & N=3 of 200 neurons                     & 2134       & 20          & 2000      & 10000  \\
Autoencoder $\tau$   & 1e-3 & 1e-6 & N=3 of 200 neurons                     & 6402       & 20          & 2000      & 10000  \\
SPNN                 & 1e-3 & 1e-5 & N= 13 of 195 neurons                   & 13         & 195         & 1000      & 5000   \\
GRU                  & 1e-3 & 1e-5 & 3 GRU of 26 neurons + 1 FC layer of 13 & 16 x 42    & 13          & -         & 10000  \\ \hline
\end{tabular}
}
\caption{Training parameters of the source model}
\label{table_param}
\end{table*}

\section{Results}

\subsection{Cognitive digital twins in a purely computational scenario}

The employ of pseudo-experimental data coming from simulations allows us to compute precise error measurements on the different adaptation strategies and their effect on diverse quantities of interest. Given a learned simulator trained off-line for glycerine data, we first apply the update methodology to learn three different liquids presenting rather diverse behaviors: water and butter as Newtonian fluids, and blood, simulated as a non-Newtonian liquid. We employ synthetic data consisting in snapshots taken from four simulations for each liquid performed under different velocity conditions to trigger diverse sloshing results. From this information, we prepare the dataset with sequences of the positions of the particles that belong to the free surface. I.e, we track the free-surface particles, take their position, and prepare sequences of 16 snapshots for each time step $\Delta t$ (the time instant of interest and the 15 previous states of the free surface). The length of 16 was found to be the minimum length of the sequences to find an embedding of the free surface measurements on each fluid's latent manifold, already computed off-line. The dataset of water has 750 snapshots in total, and butter and blood, 480. Synthetic data is available at a sampling frequency of $200 Hz$, which stands for a time step of $\Delta t = 0.005$ seconds. Therefore, the source model was built for this availability of data and time step. Nevertheless, the snapshots are sampled at every frequency of $60 Hz$, or equivalently $\Delta t= 0.015$ seconds, that matches the performance frequency of the commodity camera that will be employed in the real scenario. Hence, the change in the time step has also an effect on the model that should be corrected. The sequences of the dataset are randomly split into two subsets: 80\% for training and 20\% for testing. 

\begin{figure}[h!]
\centering
\includegraphics[width=0.7\columnwidth]{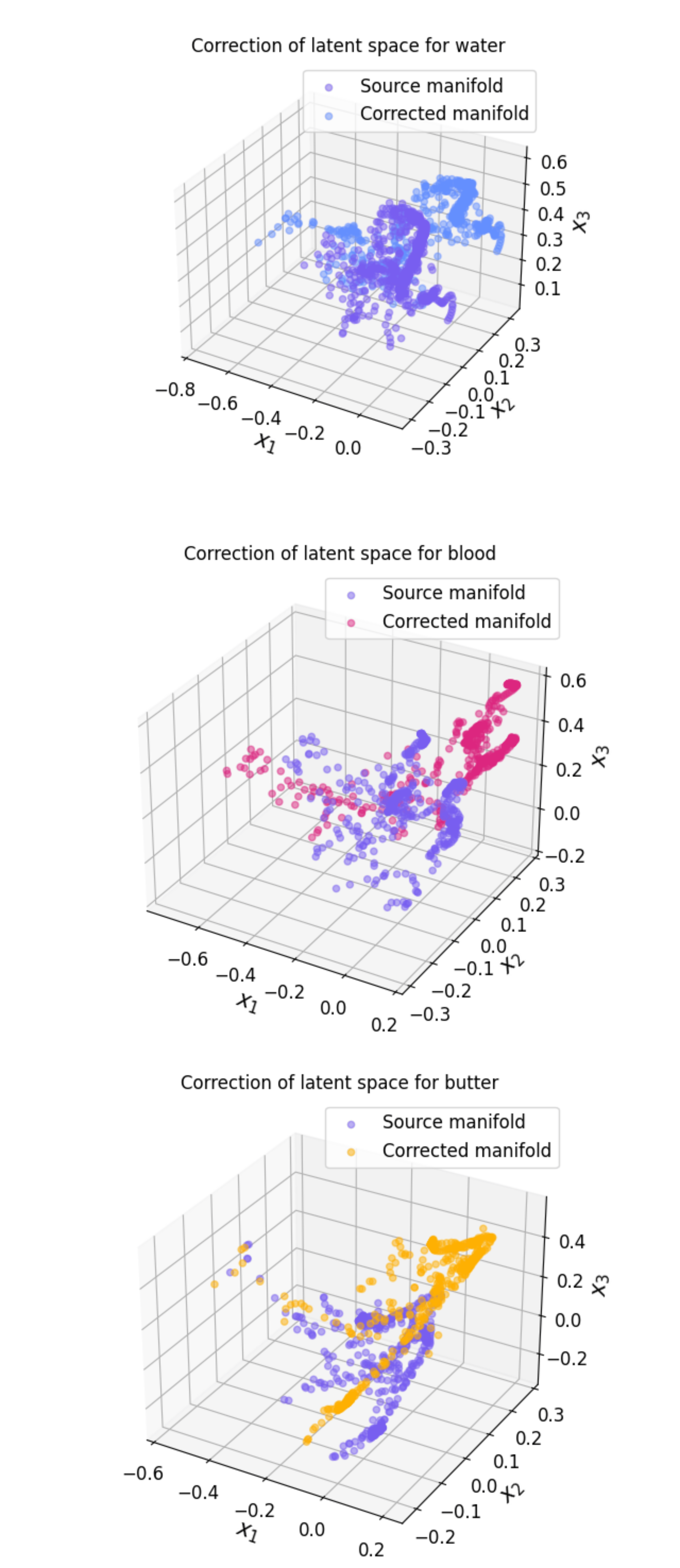}
\caption{Representation of the correction of the latent manifold of the position. The latent representation evolves to match the features of the new liquid.}
\label{fig4}
\end{figure}

We carry out the correction in the integration scheme and the embedding onto a lower-dimensional manifold. The correction is accomplished by activating the backpropagation in the last layer of the GRU network, and the 4 last layers, out of 13, of the SPNN. Since those are precedent layers of the decoder, and we train the network as a whole, we achieve the desired reconstruction with no need of altering the last structure of the network. The adaptation converges after 2000 epochs, at a small learning rate $ lr=0.0005$ and weight decay $wd=0.00001$. We choose Adam optimizer \cite{kingma2014adam}. The reconstruction is weighted by a factor $\lambda=2000$. Fig.\ref{fig4} shows the transition from the original manifold, trained off-line, to the new latent space that fits the emulated dynamics. Particularly, we show the three first state variables of the latent vector $\bs z$, that correspond to the latent variables of the position. Since the liquids bear a resemblance with the original liquid, glycerine, their manifolds do not show drastic changes in their structure compared to the initial solution. This fact also highlights the generality of the patterns of the dynamics learned in the source glycerine simulator.

The height reference is placed at the bottom of the glass and, therefore, the relative error is computed with the total height of the points of the free surface selected for the evaluation:
$$
\text{error}=\frac{1}{N}\sqrt{\sum_{n=1}^{N}\frac{\bs z_{n}^2-\hat{\bs z}_{n} ^2}{\bs z_{n}^2}},
$$
with $N$ the number of samples in the dataset, and $\bs z_{n}$ and $\hat{\bs z}_{n}$ the ground truth and simulated free surface data, respectively.

\begin{figure*}[!t]
\centering
\subfloat{\includegraphics[width=\linewidth]{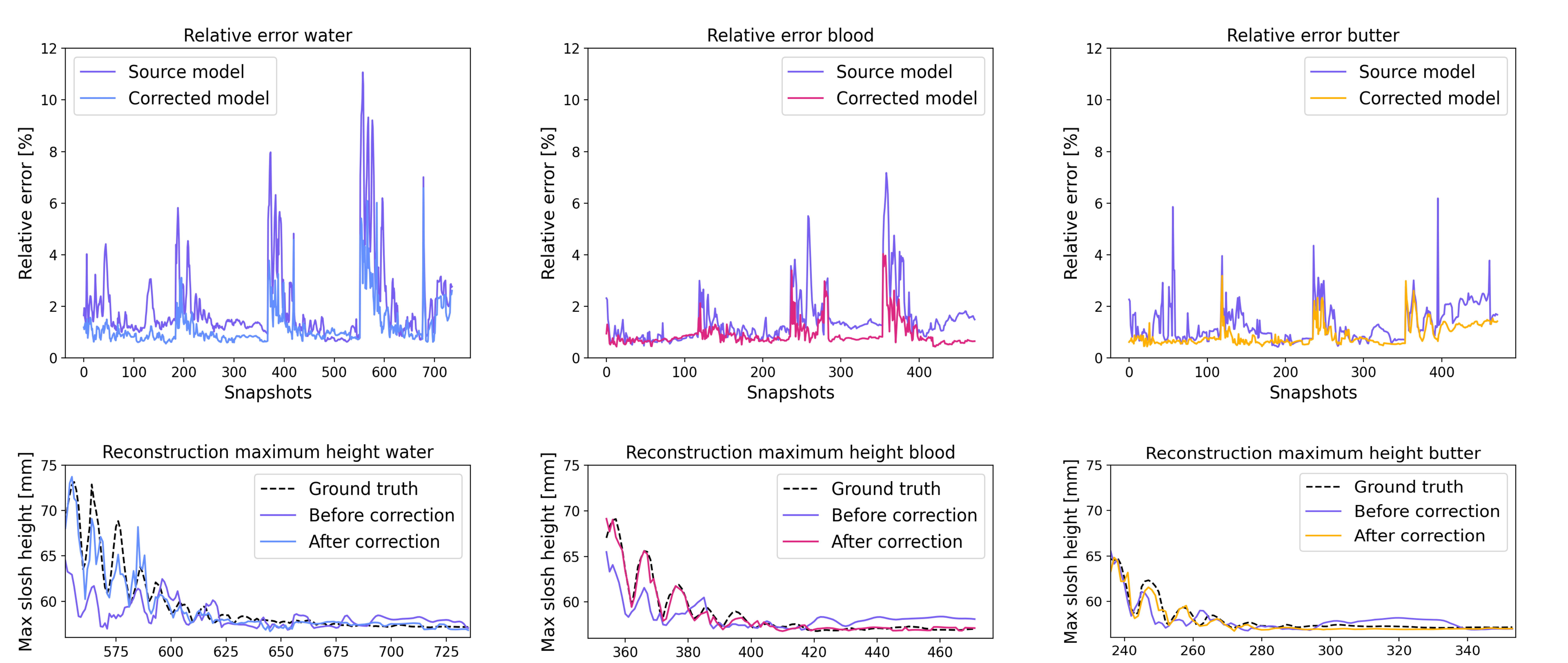}}
\hfil
\caption{(Top) Relative error of the reconstruction of the free surface for the four simulations used for training and test. The relative error obtained with the source model is reduced after the optimization. (Bottom) Detail of the sloshing height reconstruction of the most critical simulation. The method correctly emulates the behavior not only in magnitude terms but, notably, in the precise time occurrence of the peaks.}
\label{fig5}
\end{figure*}

The relative error is represented in Fig. \ref{fig5}. The update algorithm improves the performance of the simulator, reducing significantly the reconstruction error of the free surface. The relative error is also presented for the complete dataset of each liquid, reconstructing the $80\%$ employed in the correction, and the $20\%$ considered for test purposes. The four simulations can be distinguished in the graph. The error peaks appear in the correspondent top peaks of the sloshing of each simulation, where the surface deformation is greater. Nevertheless, the error continues to be in low ranges. 

Water presents higher sloshing than glycerine. Thus, the desired solutions were out of the source database. In this case, the maximum error drops from up to $12 \%$ to $4-6\%$, remaining at an average of not more than $3\%$. Butter reconstruction error is reduced from $6\%$ to less than $2\%$. Finally, the algorithm manages accurately to evolve from a Newtonian to a non-Newtonian fluid to replicate the behavior of blood. For this liquid, the maximum relative error goes from $7\%$ to $3.5\%$. It is worth mentioning that, providing that we employ a meshless method (SPH) to generate synthetic data, we can observe some errors coming from the different particle distributions. Although the movement matches the ground truth, we can have two different particle configurations. This results in different interpolated free surfaces that cause little errors despite their strong resemblance.

\begin{figure}[h!]
\centering
\includegraphics[width=0.9\columnwidth]{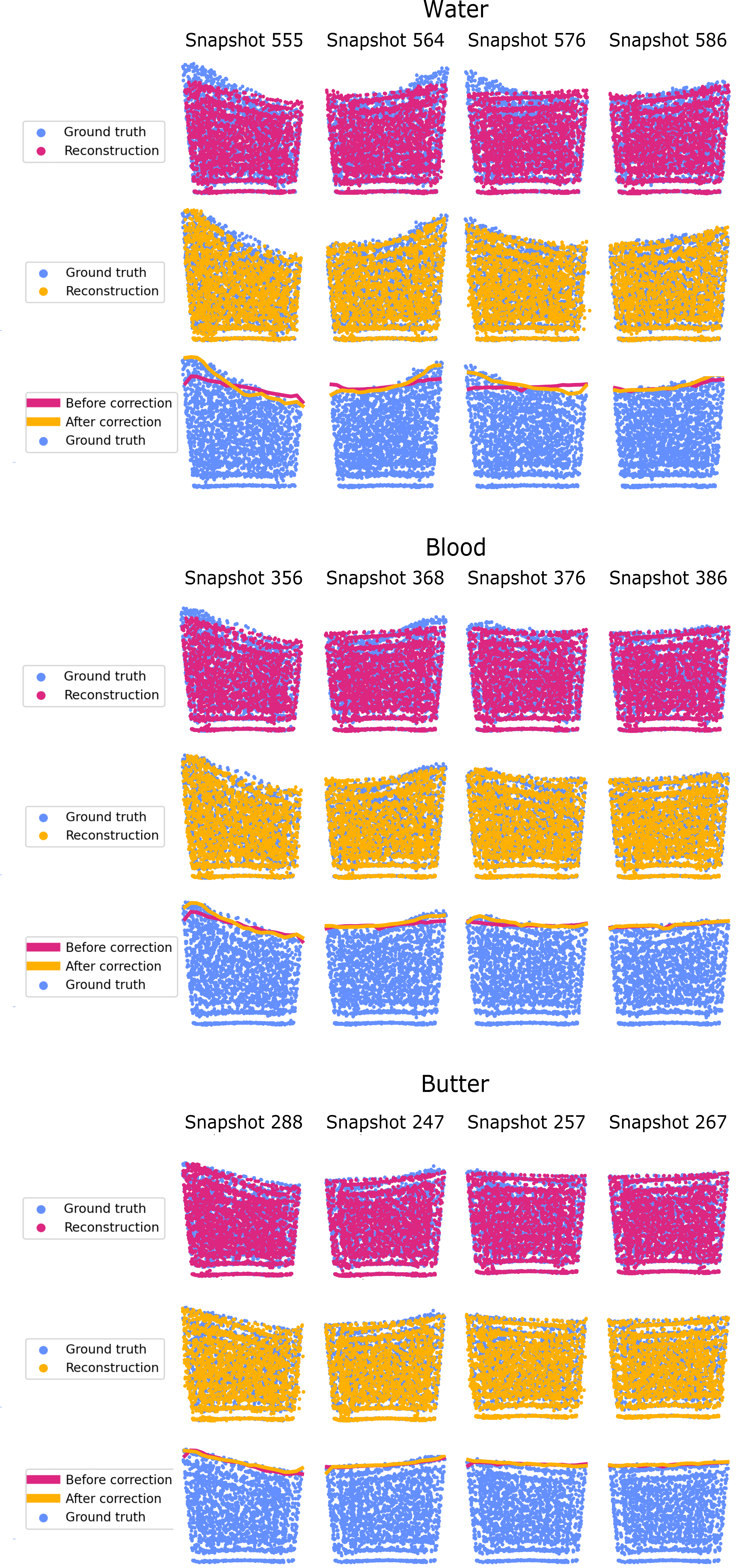}
\caption{Fluid reconstruction before and after correction of water blood and butter. Representation of the sloshing peaks of the ground truth (in blue) compared to the network’s results before (in red) and after (in yellow) applying correction to adapt from observations in selected critical snapshots, indicated in the figure.}
\label{fig6}
\end{figure}

We analyze the performance of the correction by evaluating the maximum height of the slosh in the most critical simulation for each liquid in Fig.\ref{fig6} This detail is correlated with the correspondent segment of the relative error graphs. Finally, it is worth highlighting that the reconstruction not only adjusts better to the magnitude of the slosh but also their occurrence in time. Before the correction, some peaks appear delayed or are skipped.

\subsection{Adaptation to a real-world scenario}

Employing pseudo-experimental data coming from simulations has the great advantage of allowing the computation of precise error measurements. However, the ultimate objective of this work is to develop a methodology for real-life scenarios. In this scenario, the algorithm must adapt to previously unseen real liquids with different properties that result in different frequencies and magnitudes of slosh and dissipation and duration of slosh. This correction will be done from the evaluation of the free surface of liquids tracked utilizing computer vision. This problem showcases various challenges such as errors in depth estimation, the difficulty to detect and track the free surface, the complexity of real liquids to capture all the dynamical features from the recorded videos, and user actuation. The last statement refers to the direction of the movement of the liquid. In this work, we limit the experiment to a plane movement in two possible directions (move the glass to the left or the right), but we can experience slight deviations in the actuation. We impose this restriction to bound the problem to a case where we can evaluate properly the free surface from a fixed position of the camera with the detection method employed. 

Four different liquids are evaluated in the updating framework here proposed: water, honey, beer, and gazpacho (traditional Spanish cold soup). These four materials cover a wide spectrum of behaviors and textures, from lower to higher viscosities than glycerine. In addition, the liquids selected are daily products of interest in real manipulation, that can show different behaviors depending on the production process. The glass employed is of the same shape as the computational container employed in the simulations of the source model, and the vessel is filled approximately to the same height. 

There are two recordings for each selected liquid. The first video, divided into $80\%$ snapshots for training and $20\%$ for validation to prevent overfitting, is employed in the update algorithm. The second video evaluates the performance of the new models over unknown datasets. Given these data, we apply the correction scheme for each liquid individually, obtaining four new simulators, adapted for each scenario. Equivalently to the computational test case, the correction is performed in selected active layers. Considering the complexity of the liquids' behavior and the noise of the samples, the last layer of the recurrent neural network (GRU) and the 5 last layers, out of 13, of the SPNN are activated for the correction. The correction of the recurrent neural network will adapt the manifold to the new liquid, while the partial activation of the SPNN will adapt the simulation to the new dynamics observed. The real benchmarks require more correction than the computational benchmarks. However, we maintain low learning rates and partial activation to prevent the network from learning the measurements' noise and forgetting the previously learned dynamical patterns. The correction converges after 4000 epochs, at a small learning rate $ lr=0.0001$ and weight decay $wd=0.00001$. The reconstruction is weighted by a factor $\lambda=2000$.

\begin{figure*}[!t]
\centering
\subfloat{\includegraphics[width=\linewidth]{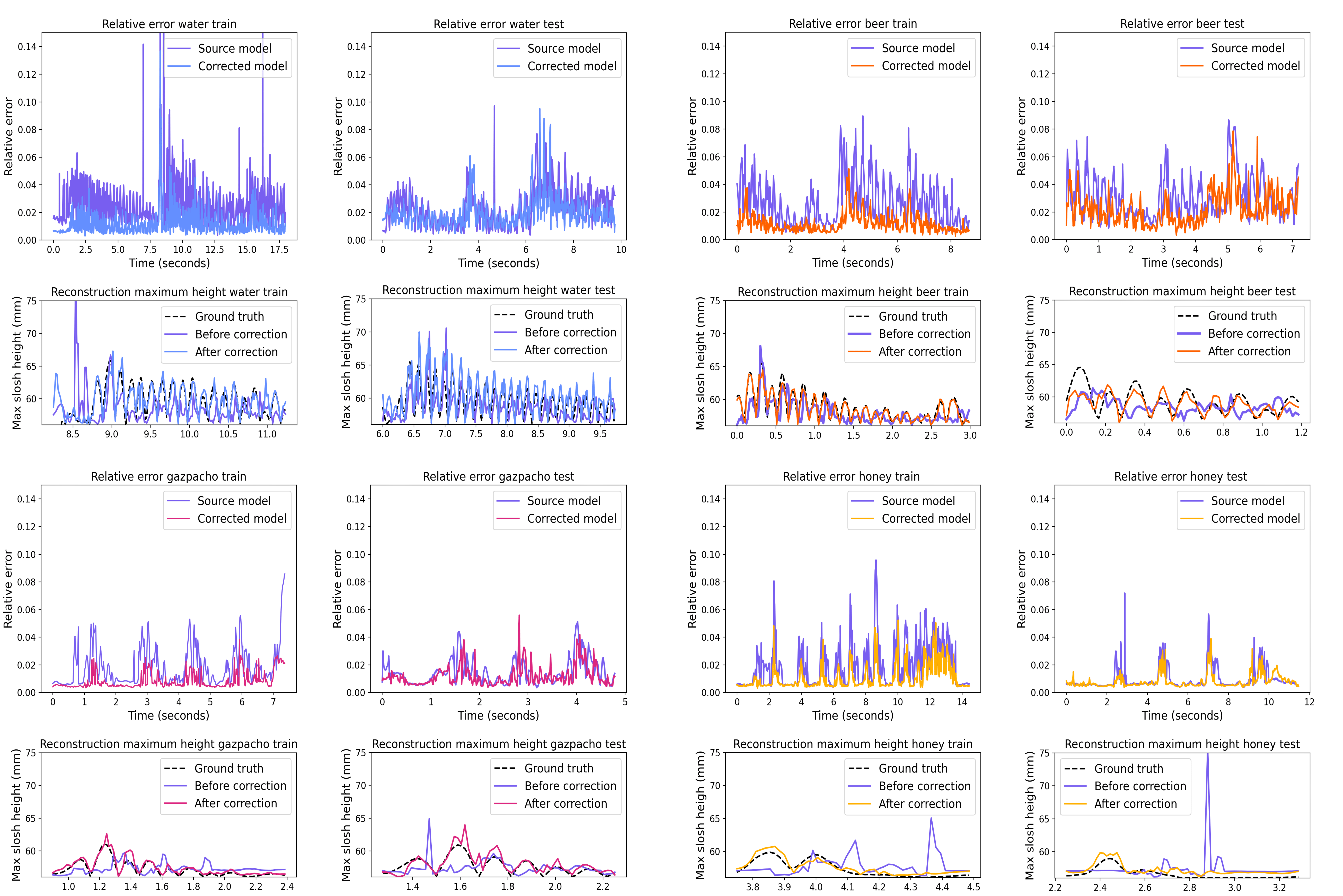}}
\hfil
\caption{Results of the adaptation to new computational liquid for water beer gazpacho and honey. The relative errors of the reconstruction are presented for the training database and over the information of an unseen recording. Despite the complexity and the few data employed, the correction in the test dataset is also noticeable. Relative error representations are accompanied by the maximum heigh reconstruction of one of the sloshes perceived to compare the performance of before and after correction.}
\label{fig7}
\end{figure*}

Fig. \ref{fig7} showcases the results of the correction over training and testing recordings of the four liquids. The four exhibit an improvement of the sloshing reconstruction compared to the performance of the source model before correction. In addition, the temporal integration with data from the test datasets presents a noteworthy performance considering that this information is new to the network. The training recordings are shorter than 10 seconds, and only three or four sloshes are captured in these datasets. However, the new model learns the new target behavior. Even though we work in a low-data regime to perform the correction of complex liquids, the method successfully reproduces the train and test benchmarks. These results are obtained because of the inductive biases learned and preserved in the correction, leading to an efficient correction from limited data and partial measurements, in this case only considering the free surface for the reconstruction. 

Water and gazpacho present slightly higher errors than the other liquids in the test dataset. This is probably due to the difficulties experienced in the data acquisition and noise of the samples in the experiment. Additionally, water has higher slosh frequencies and longer slosh time. Therefore, more varied sloshes would be needed to reduce the error, still in a low-data regime. Despite this, the method already performs the adaptation over train and test information correctly.

\begin{figure*}[!t]
\centering
\subfloat{\includegraphics[width=\linewidth]{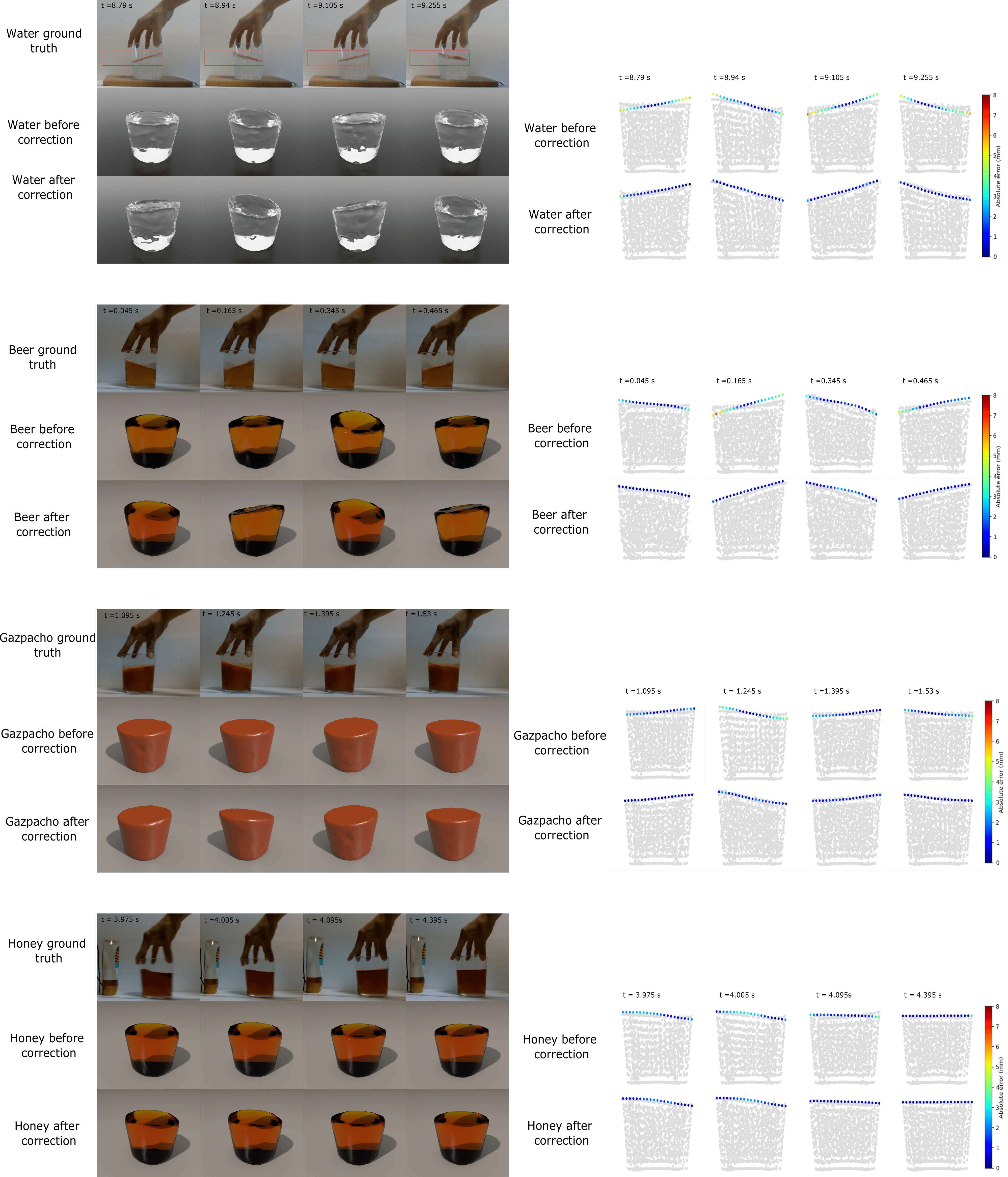}}
\hfil
\caption{Render of the volume reconstruction before and after correction. The algorithm outputs a particle discretization of the fluid that can be presented as a three-dimensional render of the volume for visualization and interpretation. The figure shows the peaks of the dynamics observed in a piece of the recordings.}
\label{fig8}
\end{figure*}

Finally, Fig. \ref{fig8} shows renders of the simulation results before and after correction compared to the snapshot that they predict in time. Additionally, the renders are complemented by a comparison between the particle discretized solution and the  ground truth's free surface. The absolute error is indicated in mm in a colored scale to analyze visually the magnitude of the distortions more accurately. It is worth mentioning that, although the particle configuration could be distorted after training to match the free surface (since it is the only information available) losing the sense of the geometry, the method manages to preserve the fluid volume configuration. The time is indicated to correlate each render to the error in Fig. \ref{fig8}. Despite the correction, the algorithm outputs a matching shape with the real entity. As a result, giving only the free surface of the observed fluids, the networks integrate in time the dynamical evolution of the liquid and provide a three-dimensional reconstruction of the fluid volume. The particle discretization represents a fluid volume that can be translated to the user through renderization and augmented reality. This representation bridges the gap between reality and the virtual environment to provide augmented information to the machine, and the user, for decision making.

\section{Conclusion}

The present work shows a DDDAs methodology for cognitive digital twins guided by GENERIC as an inductive bias for perception and reasoning about fluid sloshing. The algorithm learns from observations to  accurately mimic new fluid behaviors from the sole observation of the free surface. The method provides a tool for model inference with real data from partial observations of complex dynamics. The correction of physics perception enables the machine to adapt and learn previously unseen liquids present in daily tasks with unknown properties. We start from a source model trained with computational data to learn a physically sound simulator of the sloshing dynamics upon GENERIC to ensure the physical consistency and generalization of the results. The calculation is performed in a low-dimensional manifold to ensure real-time performance. Thus, we obtain real-time interaction with the environment in which the model, or digital twin of the real liquid, operates to have response capacity. 

We illustrate the benefits of physics-informed deep learning for correction. Given an off-line learned simulator for one particular fluid (glycerine, in our case), our method manages to evolve to a new representation of the dynamics to match the behavior of previously unseen liquids. The recordings employed are limited to 10 seconds approximately. In this low-data regime, the method adapts to the new dynamics perceived. The good performance of the method can be observed also in the test recording, which has not been seen by the network before. The success is attributed to the insights learned in the source model with simulation data, and the inductive bias imposed by GENERIC, that ensures the fulfillment of the principles of thermodynamics. These are sufficiently general and precise to allow the simulator to evolve smoothly to new liquids.Table \ref{table_update} summarizes the parameters used in the correction step. 

\begin{table*}[]
{ 
\begin{tabular}{@{}lclclclclclcl@{}}
\hline
              & lr   & wd   & unfrozen layers in GRU selected/total & unfrozen layers in SPNN selected/total & $\lambda$ & epochs \\ \hline
Comp. water   & 5e-5 & 1e-5 & 1/3                                   & 4/13                                   & 2000      & 2000   \\
Comp. blood   & 5e-5 & 1e-5 & 1/3                                   & 4/13                                   & 2000      & 2000   \\
Comp. butter  & 5e-5 & 1e-5 & 1/3                                   & 4/13                                   & 2000      & 2000   \\
Real water    & 1e-4 & 1e-5 & 1/3                                   & 5/13                                   & 2000      & 4000   \\
Real beer     & 1e-4 & 1e-5 & 1/3                                   & 5/13                                   & 2000      & 4000   \\
Real gazpacho & 1e-4 & 1e-5 & 1/3                                   & 5/13                                   & 2000      & 4000   \\
Real honey    & 1e-4 & 1e-5 & 1/3                                   & 5/13                                   & 2000      & 4000   \\ \hline
\end{tabular}
}
\caption{Update parameters}
\label{table_update}
\end{table*}

A challenge in physics perception is the balance between adaptivity and the risk of learning noise coming from the experimental nature of the data acquisition technique. Liquids and vessels are non-Lambertian, i.e., they do not have a diffusely reflecting surface, or matter, convenient for depth estimation. Despite the fine-tuning of the camera, the measurements include noise and invalid measurements from which the free surface has to be reconstructed. By applying transfer learning and performing slow training we have prevented the network from learning meaningless information coming from measurements. Moreover, the patterns already learned help reconstruct the information to learn the new behaviors accurately. 

Despite the accuracy observed in the reconstruction of train and test recordings, the performance of the method could be further improved over the last model learned by retraining with new datasets acquired by the same means. Hence, it can perform corrections to not only persist in the improvement of the reconstruction but also to adapt to the evolving nature of the scenario. 

The results observed can be a starting point to adapt the method to new geometries and include this new parameter in the optimization from the application of geometric deep learning. In addition, this problem could be further extended with more general liquid detection techniques \cite{sajjan2020clear} \cite{do2016probabilistic} capable of analyzing transparent and not textured elements from different perspectives.

  \section*{Acknowledgments}

This work has been partially funded by the Spanish Ministry of Science and Innovation, AEI /10.13039/501100011033, through Grant number PID2020-113463RB-C31 and the Regional Government of Aragon and the European Social Fund, group T24-20R. The authors also acknowledge the support of ESI Group through the project UZ-2019-0060.

\bibliographystyle{abbrv}

\end{document}